\documentclass[preprint,12pt]{elsarticle}

\usepackage{cite}
\usepackage{graphicx}
\usepackage{geometry}
\usepackage{booktabs}
\usepackage{amsfonts}
\usepackage{amsmath}
\usepackage{float}
\usepackage{subcaption}

\journal{Neural Networks}

\graphicspath{ {images/} }

\begin{document}

\begin{frontmatter}

\title{Locally Connected Spiking Neural Networks\\ for Unsupervised Feature Learning}

\author[a]{Daniel J. Saunders\corref{cor1}}
\ead{djsaunde@cs.umass.edu}

\author[a]{Devdhar Patel}
\ead{devdharpatel@cs.umass.edu}

\author[a]{Hananel Hazan}
\ead{hhazan@cs.umass.edu}

\author[a]{Hava T. Siegelmann}
\ead{hava@cs.umass.edu}

\author[a,b]{Robert Kozma\corref{cor1}}
\ead{rkozma@cs.umass.edu}


\address[a]{Biologically Inspired Neural and Dynamical Systems Laboratory (BINDS)\\
University of Massachusetts Amherst, College of Computer and Information Sciences\\
140 Governors Drive, Amherst, MA 01003, USA}

\address[b]{Center for Large-Scale Intelligent Optimization and Networks (CLION)\\
Department of Mathematics, University of Memphis, Memphis, TN 38152, USA}

\begin{abstract}

In recent years, Spiking Neural Networks (SNNs) have demonstrated great successes in completing various Machine Learning tasks. We introduce a method for learning image features by \textit{locally connected layers} in SNNs using spike-timing-dependent plasticity (STDP) rule. In our approach, sub-networks compete via competitive inhibitory interactions to learn features from different locations of the input space. These \textit{Locally-Connected SNNs} (LC-SNNs) manifest key topological features of the spatial interaction of biological neurons.
We explore biologically inspired n-gram classification approach allowing parallel processing over various patches of the the image space.
We report the classification accuracy of simple two-layer LC-SNNs on two image datasets, which match the state-of-art performance and are the first results to date. LC-SNNs have the advantage of fast convergence to a dataset representation, and they require fewer learnable parameters than other SNN approaches with unsupervised learning. Robustness tests demonstrate that LC-SNNs exhibit graceful degradation of performance despite the random deletion of large amounts of synapses and neurons.

\end{abstract}

\cortext[cor1]{Corresponding author}
\end{frontmatter}

\section{Introduction}
\label{intro}

Processing of image data by machine learning (ML) systems often requires techniques for overcoming the problem known as the \textit{curse of dimensionality}; that is, the inherent difficulty in processing high-dimensional data. The locally-connected layer of the neural networks literature \citep{Chen2015LocallyconnectedAC} imposes an inductive bias on an artificial neural network (ANN) that stipulates that certain features only occur in particular regions of input space. This is in sharp contrast to the popular \textit{convolutional layer}, in which a set of filters are learned that can be used to detect features across the entire input space, inducing a \textit{translation invariant} data representation \citep{KaudererAbrams2016QuantifyingTI}; i.e., in which the occurrence or relative locations of features are important. In the deep learning (DL) literature, both methods were key to reducing the number of parameters needed to train deep and relatively efficient networks for processing complex data; e.g., images and videos [citations needed].

Spiking neural networks (SNNs) \citep{MAASS19971659} are well-known for their energy efficiency and speed on neuromorphic hardware platforms \citep{DBLP:journals/corr/abs-1710-04838}. For this reason, development of methods that enable SNNs to reach the performance of ANNs on machine learning tasks has gained interest in recent years \citep{deeplearninginsnns}. Previous work has shown that spiking neural networks can, in principle, be used for performing complex machine learning tasks; e.g., image classification \citep{diehl_fast-classifying_2015, DBLP:journals/corr/abs-1802-02627} and reinforcement learning \citep{6796089, Mozafari2018FirstspikeBV}. On the other hand, few methods exist for the robust training of SNNs from scratch for general-purpose machine learning ability; i.e., their abilities are highly domain- or dataset-specific, and require much data pre-processing and tweaking of hyper-parameters in order to attain good performance.

This paper aims to extend the abilities of spiking neural networks by introducing and studying a spiking version of a computational layer borrowed from the artificial neural networks literature. In particular, the \textit{locally connected layer} is introduced and implemented in spiking neural networks, to which arbitrary biological learning rules (e.g., Hebbian learning, spike-timing-dependent plasticity, etc.) may be applied \citep{10.1371/journal.pcbi.1006227}. Simple two-layer networks are built and trained in an unsupervised fashion to produce a representation of the MNIST \citep{lecun-mnisthandwrittendigit-2010} and EMNIST \citep{2017arXiv170205373C} datasets, and ultimately, the goodness of the representation (i.e., to what extent the learned features linearly separate the classes of data) is assessed through a simple \textit{voting mechanism} on the spiking outputs of the trained networks. In particular, we use the $n$-gram method of \citep{Hazan2018UnsupervisedLW}.
The introduced spiking network component and corresponding training procedure produce distributed representations which are robust to a large degree of randomly deleted synapses or neurons, a property commonly referred to as \textit{graceful degradation}.


There are various popular libraries in the literature that allow efficient spiking neural network simulations. Some of them provide deep elaboration of biological details; see,
\texttt{ANNarchy} \citep{Vitay2015ANNarchyAC},
\texttt{BRIAN} \citep{goodman_brian_2009},
\texttt{Genesis} \citep{GENESIS3.0},
\texttt{NEURON} \citep{NEURON}, and
\texttt{NEST} \citep{Gewaltig_NEST}.
Other approaches focus on high-level behaviors of spiking
neural networks, with benefit for machine learning applications, e.g.,
\texttt{NeuCube} \citep{KASABOV201462}, and
\texttt{NENGO} \citep{bekolay2014}.
We developed the spiking neural network software package \texttt{BindsNET} written in Python \citep{hazanetal2018}. \texttt{BindsNET} provides us with significant freedom in deciding implementation details, aiming at machine learning applications.
\texttt{BindsNET} builds on the results of \citep{saundersetal} and \citep{p._u._diehl_unsupervised_2015}, wherein all networks were implemented in the \texttt{BRIAN} spiking neural networks simulator \citep{d._f._m._goodman_brian_2009}. In addition to image classification tasks addressed here, \texttt{BindsNET} has been used implementing SNNs playing computer games such as Atari breakout, by transferring weights from Deep Q-Learning NNs trained by reinforcement learning \citep{Dedvhar19}.
The SNNs introduced in this paper are implemented in \texttt{BindsNET} and have been trained by STDP algorithm to solve image recognition tasks.
The \texttt{BindsNET} library enables the simulation of networks on CPUs or GPUs or a combination thereof, and GPU computation is useful in particular for the fast simulation of large LC-SNNs with many neurons and synapses.

\section{Related work}
\label{related_work}

Due to their potential for efficient implementation on neuromorphic hardware devices, much recent work has focused on developing powerful and efficient learning algorithms for SNNs. We focus on SNNs trained for machine learning tasks in simulated environments, which may be either (1) converted to neuromorphic platforms post-training, or, in some cases (2) trained and evaluated entirely on neuromorphic chips. A review of machine learning with SNNs is treated by \citep{deeplearninginsnns}.

This work is related most closely to that of Diehl \& Cook \citep{p._u._diehl_unsupervised_2015}, in which a simple three-layer network is trained unsupervised with spike-timing-dependent plasticity along with excitatory-inhibitory interactions between neurons to learn to classify the MNIST handwritten digits \citep{lecun-mnisthandwrittendigit-2010}. The input layer is comprised of Poisson spiking neurons fully-connected with STDP-modifiable synapses to an arbitrarily-sized layer of excitatory neurons. Neurons in this layer ``compete'' with one another to learn filters that capture prototypical examples from the data via connections from an inhibitory layer. Once trained, this network operates similarly to a \textit{winner-take-all} (WTA) circuit, with sparse activity in the excitatory population coding for a single category of data. Several extensions of this work has been explored quite recently, including the simultaneous clustering and classification of image data \citep{Hazan2018UnsupervisedLW} and the incremental learning of distinct categories \citep{Allred2016UnsupervisedIS}.

Several other unsupervised learning methods have been developed and demonstrated with SNNs. A network inspired by the autoencoder of the neural networks literature is trained without labels layer-wise to reconstruct the MNIST and CIFAR-10 datasets, and whose output is trained in a supervised fashion to perform classification \citep{DBLP:journals/corr/PandaR16}. A learning algorithm for SNNs is developed on the supposition that neurons tend to maximize their activity in competition with other neurons in \citep{10.3389/fninf.2018.00079}. Three-layer networks of Izhikevich regular spiking neurons are trained with STDP on the binary task of distinguishing the digits 0 and 1 as well as on the full set of MNIST digits \citep{Tavanaei2015AMS}.

In this paper, we are primarily interested in the implementation of locally-connected layer (see, e.g., \citep{Chen2015LocallyconnectedAC}) for spiking neural networks. A closely related concept is the convolutional layer \citep{Goodfellow-et-al-2016}, which is named for how it convolves a number of weight kernels across the input space while the weights remain fixed, or \textit{shared} between locations in the input space. A number of studies have implemented convolutional layers for SNNs, with a mixture of supervised and unsupervised learning methods \citep{Tavanaei2017MultilayerUL, Lee2018TrainingDS, Tavanaei2018TrainingSC, Tavanaei2016BioInspiredSC, KHERADPISHEH201856}. Our locally connected layer is similar to this concept, except that a different set of weight kernels is learned per input location; i.e., the weights are not shared as we stride across the input space.


\section{Methods}
\label{sec:methods}

\subsection{Spiking neuron model}

Our networks use a variant of the popular \textit{leaky integrate-and-fire} (LIF) spiking neuron with adaptive firing thresholds. The dynamics of the LIF voltage $v(t)$ is given by

\begin{equation}
    \tau_v \frac{\textrm{d} v(t)}{\textrm{d} t} = -v(t) + v_\textrm{rest} + I(t),
\end{equation}

where $\tau_v$ is the membrane time constant, $v_\textrm{rest}$ is the neuron's \textit{resting voltage} (to which it decays exponentially in time), and $I(t)$ denotes the total input to the neuron at time $t$. The voltage (membrane potential) $v(t)$ of a \textit{post-synaptic} neuron is increased at the occurrence of a spike from a \textit{pre-synaptic} neuron connected to it by the weight $w$ of the synapse that connects them; all such incoming currents are gathered in $I(t)$. When a neuron reaches or exceeds its threshold membrane potential $v_\textrm{thresh} (t)$, it emits a spike to downstream neurons and resets to a voltage $v_\textrm{reset}$. At this point, the neuron is clamped to the reset voltage for a short \textit{refractory period} $\Delta_\textrm{ref}$ and does not integrate its inputs.

The adaptive threshold $\theta(t)$ is used to ensure that a single neuron does not end up dominating the firing activity of the output layer, and that different neurons tune selectively to the intensity of different prototypical inputs. It has dynamics

\begin{equation}
    \tau_\theta \frac{\textrm{d} \theta (t)}{\textrm{d} t} = -\theta (t),
\end{equation}

where $v_\textrm{thresh} (t) = \theta_0 + \theta(t)$, with $\theta_0 > v_\textrm{reset}, v_\textrm{rest}$ a constant. Each time a neuron fires a spike, the adaptive threshold $\theta$ is increased by a constant $\theta_\textrm{plus}$.

In our experiments, we set $\theta_0 = -52$mV, $v_\textrm{rest} = v_\textrm{reset} = -65$mV, $\Delta_\textrm{ref}$ = 5ms, $\tau_v$ = 20ms, $\tau_\theta$ = 1000s, and $\theta_\textrm{plus}$ = 0.05. In the context of machine learning, it may be of interest to include these in a grid- or random-search hyper-parameter optimization algorithm, but for the experiments described in this paper, they remain fixed.

Input neurons (used to convert non-negative image data to spikes in our experiments) are modeled by \textit{Poisson spikes trains} with average firing rate equal to the input intensity in Hertz. For the experiments in Section \ref{sec:results}, we multiply the MNIST and EMNIST images (having grayscale values in [0, 255]) by a factor of $\frac{1}{2}$ to constrain input firing rates from 0Hz to 127.5Hz.

\subsection{Learning rule}
\label{ssec:learning}

Our locally-connected SNN layers are trained with a simple spike-timing-dependent plasticity (STDP) rule combined with a weight normalization scheme. The STDP rule combines both pre- and post-synaptic weight updates in order to detect temporally coincidental spikes from connected neurons. Our STDP rule has the form:

\begin{equation}
    \Delta w =
    \begin{cases}
        \eta_\textrm{post} x_\textrm{pre} & \textrm{on post-synaptic spike;} \\
        -\eta_\textrm{pre} x_\textrm{post} & \textrm{on pre-synaptic spike.}
    \end{cases}
\end{equation}

The parameters $\eta_\textrm{pre}, \eta_\textrm{post}$ are pre-, post-synaptic learning rates, respectively, and the quantities $x_\textrm{pre}, x_\textrm{post}$ are pre-, post-synaptic \textit{spike traces}, which are set to 1 on (pre-, post-synaptic) spike events, and are exponentially decaying to 0 otherwise. The spike traces are a simple memory trace of a neuron's previously-occurring spike, used for implementing STDP in an \textit{online} fashion; i.e., carried out during the SNN simulation. In our experiments, we set $\eta_\textrm{post} >> \eta_\textrm{pre}$ to emphasize the effect that pre-synaptic spikes have on post-synaptic neuron membrane potentiation. In particular, we used values around $\eta_\textrm{post} = 1 \times 10^{-2}$ and $\eta_\textrm{pre} = 1 \times 10^{-4}$, but these (and their exponential decay rate) were subject to grid search in all experiments.

Due to the unbalanced nature of the pre- and post-synaptic STDP updates, another mechansim is required to ensure that synaptic weights don't grow without bound. First and foremost, weights are clipped after each iteration to the range $[0, w_\textrm{max}]$, and secondly, the sum of weights incident to a post-synaptic neuron is normalized after each iteration to have sum $c_\textrm{norm}$ (\textit{normalization constant}). That is, if there are $n_\textrm{pre}$ pre-synaptic neurons, the synapses incident to a single post-synaptic neuron have average value ${c_\textrm{norm}}/{n_\textrm{pre}}$.

\subsection{Network architecture: locally-connected SNN}
\label{ssec:lcsnn}

The locally-connected layer is borrowed from the deep learning literature \citep{Chen2015LocallyconnectedAC} and implemented in a spiking neural network. We restrict our discussion to data with two spatial dimensions; e.g., image data, with or without multiple color channels. At its most basic, the layer requires parameters $k_1, k_2$ (horizontal and vertical kernel size, respectively), $s_1, s_2$ (horizontal and vertical stride length, respectively), and $n_\textrm{filters}$, the number of filters to learn per input location (\textit{receptive field}). Locations in the input space are determined by the kernel size and stride parameters; the first receptive field resides at the upper left corner of the input space, with spatial extent $k_1 \times k_2$, and the filter is slid over by $s_1$ pixels until the right edge of the filter is incident to the right edge of the image, at which point the filter is moved down $s_2$ pixels and back to the left edge of the image, and process starts anew. The process is repeated until the entire input space is tiled with filters.

The SNN implementation of a locally-connected layer simply stipulates that there are synapses from the pre-synaptic to post-synaptic populations which coincide with the location of filters in the pre-synaptic population, and where the size of the post-synaptic population is determined by the parameters ($k_1, k_2, s_1, s_2, n_\textrm{filters}$) of the layer. In this paper, we restrict attention to the case $k_1 = k_2$ and $s_1 = s_2$ to simplify parameter search, and refer to the parameters respectively as $k$ and $s$.

We refer to a network which contains a locally-connected layer as a \textit{locally-connected spiking neural network} (LC-SNN), although the network may contain additional layers with other layers included. Locally-connected layers are not restricted to receiving connections directly from the input; they may appear in any sequence of processing in a SNN. For example, multiple locally-connected layers can be connected in sequence to detect progressively more complex features in a hierarchical fashion.

\begin{figure}[H]
    \centering
    \includegraphics[width=0.65\textwidth]{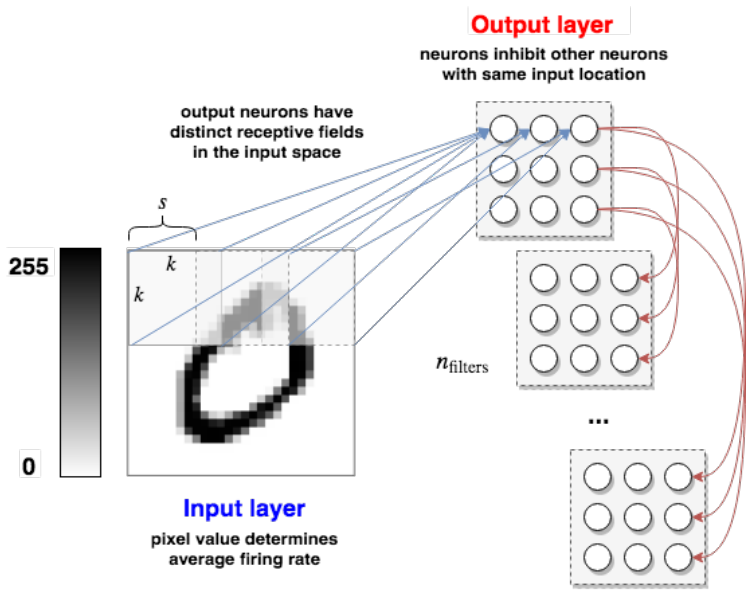}
    \caption{Two-layer locally-connected spiking neural network architecture.}
    \label{fig:lc-snn}
\end{figure}

In this work, we report classification results from a two-layer (input, output) LC-SNN. A schematic of this architecture is shown in Figure \ref{fig:lc-snn}. An input example is rate-coded as the average firing rates of Poisson spiking neurons, which are connected with locally specific, STDP-modifiable synapses to the output layer. The output layer neurons are connected to other neurons which share the same receptive field with fixed inhibitory synapses. This inhibitory connectivity forces neurons in the output to operate in a WTA-like fashion, where, ultimately, one neuron per receptive field remains firing after all others are inhibited by its activity. This is similar to the fully-connected SNN architecture of \citep{p._u._diehl_unsupervised_2015}, except that the input population is divided into a number of (possibly overlapping) sub-regions, instead of connecting the full input population to every output neuron.

\subsection{Voting mechanism}
\label{ssec:voting}

In a commonly used voting approach, each individual output spike counts as a single vote, during the observation period. The more votes a neuron gets, the more likely it is selected. In the max decision rule, the winner will be the neuron with the highest number of spikes (votes).
In this work, we explore the $n$-gram method as well to classify new input examples based on historical (training) firing activity \citep{Hazan2018UnsupervisedLW}. We set $n$ = 2 for the sake of computational expediency. Accordingly, we record all length 2 sequences of spikes from the output layer during the training phase. These sequences are aggregated over the training phase, and are paired with the class of data for which they appear the most. When a test example is presented, all length 2 sequences are recorded, and each gets a vote (based on the label they're paired with) towards the classification of the new example.

In this way, we are capturing simple information about the ordering of firing and harnessing it to distinguish the label of the data. This method is a step beyond a typical voting approach mentioned before, in which individual output spikes count as single votes. In the n-gram approach, we are taking advantage of the time domain aspect of spiking neural networks by supposing that \textit{the ordering of firing matters}; i.e., permuting the order of output spikes results in a loss of information from the perspective of network encoding.
In this work, we report only results using the $n$-gram method, but note that this method outperforms our experiments with single spike, vote-based (and re-weighted vote-based) classification schemes.

\subsection{Model selection}
\label{ssec:selection}

A problem common to many machine learning tasks is picking the best version of a trained model to use for the test phase. To this end, during the training, the test accuracy of the network is periodically estimated by evaluating it with the $n$-gram classification scheme on a random sampling of examples from the training dataset. The version of the network which obtains the highest estimated accuracy is saved to disk, to be tested on the full test dataset. In our experiments, we estimate the test accuracy after every 250 examples on a randomly sampled batch of 250 examples.

\section{Results}
\label{sec:results}

In the following sections, we present unsupervised learning results from LC-SNN networks with two-layer architectures, and show how training convergence of our methods improve over that of previous work on unsupervised learning with SNNs. These results build on those obtained by \citep{saundersetal}.

\subsection{Two-layer LC-SNN results}
\label{ssec:lcsnn_results}

We report test accuracy results from two-layer LC-SNNs with varying numbers of locally-connected filters $n_\textrm{filters}$. The network consists of an input layer of size equal to the input space (images center-cropped to 20 $\times$ 20), an output layer with size determined by parameters $(k, s, n_\textrm{filters})$, STDP-modifiable synapses from input to output layer, and a recurrent inhibitory connection in the output layer. As mentioned above, output neurons are inhibited by other output neurons that share the same receptive field in the input space.

\subsubsection{MNIST digits dataset}
\label{sssec:mnist}

Grid search is conducted to find the best setting of hyper-parameters in terms of classification accuracy, subject only to experimental time constraints. All experiments are limited to running for 24 hours with 12Gb of RAM. With large $n_\textrm{filters}$ and certain settings of $k$ and $s$, there is not enough memory or time to complete the simulation, further constraining the hyper-parameter search space. All networks are trained for 60K iterations; i.e., one pass through the MNIST training data, and tested on all 10K test examples. Both training and test examples (cropped to size 20 $\times$ 20) are presented to the network for 250ms of simulated time with a 1ms timestep. The results are reported in Table \ref{tab:mnist_results} along with the setting of kernel size ($k$) and stride ($s$) that resulted in the best accuracy per number of filters $n_\textrm{filters}$, as well as the number of STDP-modifiable parameters $n_\textrm{synapses}$ and output layer neurons $n_\textrm{neurons}$ those settings required, computed as a function of $k$, $s$, and $n_\textrm{filters}$.

\begin{table}[H]
    \centering
    \begin{tabular}{|c|c|c|c|c|}
        \toprule
        $n_\textrm{filters}$ & $(k, s)$ & $n_\textrm{synapses}$ & $n_\textrm{neurons}$ & mean $\pm$ std. \\
        \midrule
        25 & (12, 4) & 32.4K & 225 & 84.01 $\pm$ 2.28 \\
        50 & (14, 4) & 64.8K & 450 & 88.21 $\pm$ 1.79 \\
        100 & (12, 4) & 129.6K & 900 & 91.68 $\pm$ 1.31 \\
        250 & (16, 2) & 576K & 2250 & 93.71 $\pm$ 0.68 \\
        500 & (16, 2) & 1.15M & 4500 & 94.59 $\pm$ 0.61 \\
        1000 & (16, 2) & 2.3M & 9000 & \textbf{95.07} $\pm$ \textbf{0.63} \\
        \bottomrule
    \end{tabular}
    \caption{Test accuracy results for LC-SNNs with varying numbers of $n_\textrm{filters}$ using the $2$-gram classification scheme. Each result is estimated from 10 independent train and test phases.}
    \label{tab:mnist_results}
\end{table}

\begin{figure}[H]
    \centering
    \begin{subfigure}[t]{0.475\textwidth}
        \includegraphics[width=\textwidth]{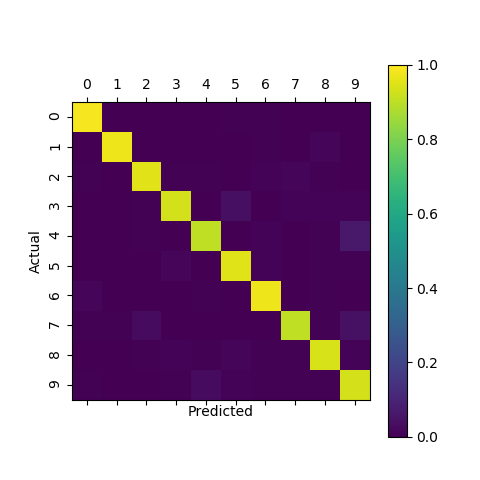}
        \caption{LC-SNN MNIST confusion matrix. The most common mistake is interchanging the digits '4' and '9'.}
        \label{subfig:mnist_confusion}
    \end{subfigure}
    \begin{subfigure}[t]{0.475\textwidth}
        \includegraphics[width=\textwidth]{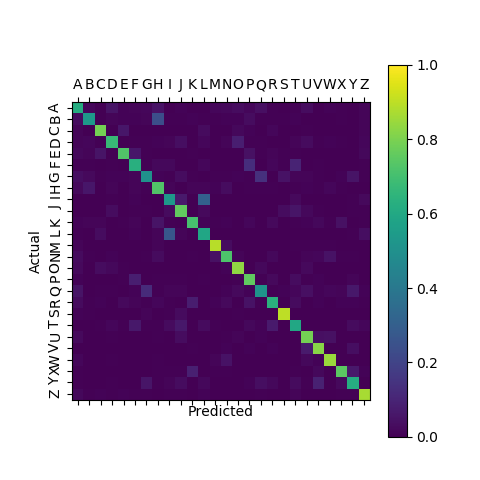}
        \caption{LC-SNN EMNIST confusion matrix. The most common mistake is interchanging the letters 'I' and 'L'.}
        \label{subfig:emnist_confusion}
    \end{subfigure}
    \caption{Average confusion matrices on the MNIST and EMNIST test partitions for 10 independently trained LC-SNNs with $n_\textrm{filters}$ = 1000, $k$ = 16, $s$ = 2.}
    \label{fig:confusion}
\end{figure}

The average test confusion matrix for the best set of hyper-parameters for LC-SNNs with $n_\textrm{filters} = 1000$ is shown in Figure \ref{subfig:mnist_confusion}.

\begin{figure}[H]
    \centering
    \begin{subfigure}[t]{0.325\textwidth}
        \includegraphics[width=\textwidth]{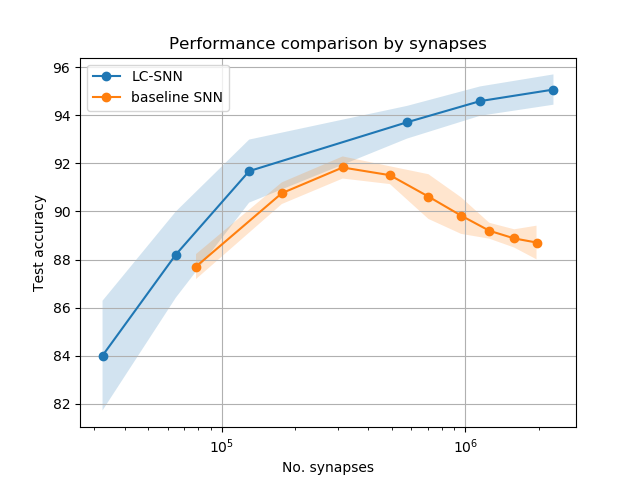}
        \caption{}
        \label{subfig:lcsnn_dac_synapses}
    \end{subfigure}
    \begin{subfigure}[t]{0.325\textwidth}
        \includegraphics[width=\textwidth]{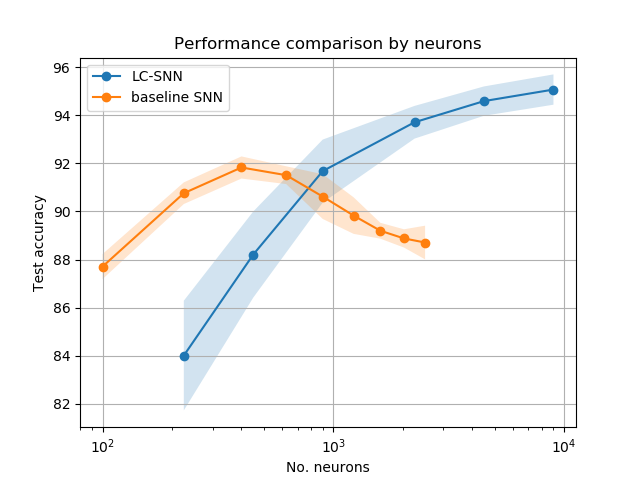}
        \caption{}
        \label{subfig:lcsnn_dac_neurons}
    \end{subfigure}
    \begin{subfigure}[t]{0.325\textwidth}
        \includegraphics[width=\textwidth]{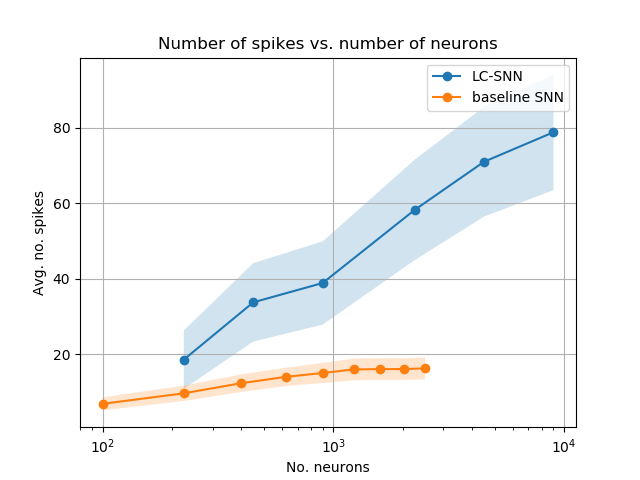}
        \caption{}
        \label{subfig:lcsnn_dac_spikes}
    \end{subfigure}
    \caption{Various tradeoffs between LC-SNNs and the baseline SNN in terms of the number of learnable synapse weights, neurons, and spikes during the test phase.}
    \label{fig:lcsnn_dac_comparison}
\end{figure}

These accuracy results are comparable to similar work by \citep{p._u._diehl_unsupervised_2015}, in which simple SNNs with a fully-connected layer are trained with STDP and competitive inhibitory interactions to classify the MNIST dataset. We refer to these networks from this point onwards as a baseline. However, our method is able to achieve higher test accuracy with fewer parameters due in part to the parameter-efficient local connectivity and improved training methods. On the other hand, the accuracy of methods is similar when compared instead by the number of neurons utilized. We plot a comparison of $n_\textrm{synapses}$ and $n_\textrm{neurons}$ versus test accuracy for the baseline SNN and the LC-SNN in Figure \ref{fig:lcsnn_dac_comparison}. By adding more locally-connected filters, we expect that LC-SNN test accuracy will exceed that of the baseline SNN's best result.

Moreover, while we limit ourselves to a single pass through the training data, the results of \citep{p._u._diehl_unsupervised_2015} are reported with 1, 3, 7, and 15 passes through the training dataset for networks with 100, 400, 1,600, and 6,400 neurons, respectively. For a better comparison, we re-implemented the baseline SNN in \texttt{BindsNET} and ran a superset of the same experiments with a single pass though the training data. These results are also depicted in Figure \ref{fig:lcsnn_dac_comparison}, from which one can see that the larger baseline SNNs are severely undertrained without multiple passes though the training data. We expect that LC-SNN performance will further improve with multiple presentations of the training set.

Figure \ref{subfig:lcsnn_dac_synapses} demonstrates that, with a single pass through the training data, any size LC-SNN (in terms of STDP-modifiable synapses) achieves a better performance than the corresponding baseline network. On the other hand, Figure \ref{subfig:lcsnn_dac_neurons} show that smaller baseline SNNs attain better accuracy in terms of the number of spiking neurons required in the output layer. However, in our simulations, the main simulation bottlenecks were related to the dimensionality of the tesnors implicated in linear algebra operations, namely by the size of the weight matrices of the SNNs. To that end, the LC-SNN demonstrates a clear advanatage over the baseline in terms of simulation training time.

Figure \ref{subfig:lcsnn_dac_spikes} shows that the LC-SNN induce many more spikes than the baseline SNN, due to the distributed nature of the recurrent inhibitory connection; i.e., whereas the baseline networks operates similarly to a winner-take-all circuit, in which a single neuron typically causes all others to fall quiescent, the LC-SNN allows one neuron \textit{per receptive field} to ``win'' in a similar manner. From a energy efficiency point of view, the LC-SNN implemented in neuromorphic hardware would be more costly than the baseline due to the increased activity, but with the benenfit of reduced training time and fewer synaptic connections. This can be seen as a tradeoff between energy and training time requirements, where the extreme sparsity of the baseline network's activity results in a bottleneck in training convergence. Note that the number of LC-SNN spikes is only $\approx$ 2-5X greater than the baseline networks' for all numbers of neurons considered.

Table \ref{tab:mnist_comparison} compares a collection of results from training spiking neural networks to classify the MNIST handwritten digits. We consider methods which explicitly use labels to learn all network parameters \textit{supervised}, whereas methods which employ a partial unsupervised training are denoted \textit{semi-supervised}. Of all considered unsupervised methods, our approach reaches state of the art results. On the other hand, semi-supervised and supervised methods are often superior, likely due to incorporating label information directly into the learning procedure.

\begin{table}[H]
    \centering
    \resizebox{\textwidth}{!}{\begin{tabular}{||c|c|c|c|c||}
        \toprule
        Paper & Architecture & Learning & mean & std. \\
        \midrule
        Ours & Local + recurrent connections & Unsupervised & 95.07 & 0.63 \\
        \citep{p._u._diehl_unsupervised_2015} & Dense + recurrent connections & Unsupervised & 95.00 & - \\
        \citep{Tavanaei2015AMS} & Input segmentation & Unsupervised & 75.93 & - \\
        \citep{Allred2016UnsupervisedIS} & Dense + recurrent connections & Incremental unsupervised & 86.59 & - \\
        \citep{DBLP:journals/corr/PandaR16} & Autoencoder & Semi-supervised & 99.08 & - \\
        \citep{10.3389/fninf.2018.00079} & Dense + lateral connections & Supervised / semi-supervised & 95.4 / 72.1 & - \\
        \citep{Tavanaei2017MultilayerUL} & Convolutional & Semi-supervised & 96.95 & 0.08 \\
        \citep{Lee2018TrainingDS} & Convolutional & Semi-supervised & 99.28 & $\approx$ 0.1 \\
        \citep{Tavanaei2018TrainingSC} & Convolutional & Semi-supervised & 97.20 & 0.07 \\
        \citep{KHERADPISHEH201856} & Convolutional & Semi-supervised & 98.4 & - \\
        \bottomrule
    \end{tabular}}
    \caption{Comparison of SNN training results on the MNIST dataset. See respective papers for details on SNN architectures and learning procedures.}
    \label{tab:mnist_comparison}
\end{table}

\subsubsection{EMNIST: letters partition}

The partition of the EMNIST dataset containing the 26 capital letters of the English alphabet is used to once again test our LC-SNN approach against that of the baseline network architecture. Grid search is once again used to select the best setting of hyper-parameters (kernel size, stride, learning rate, and learning rate decay) per setting of $n_\textrm{filters}$. A single pass through the 124K training examples is used (4.8K per class) regardless of network size. The results are reported in Table \ref{tab:emnist_results}.

\begin{table}[H]
    \centering
    \begin{tabular}{|c|c|c|c|c|}
        \toprule
        $n_\textrm{filters}$ & $(k, s)$ & mean $\pm$ std. \\
        \midrule
        25 & (12, 4) & 49.61 $\pm$ 1.98 \\
        50 & (12, 4) & 54.44 $\pm$ 2.04 \\
        100 & (12, 4) & 62.34 $\pm$ 2.11 \\
        250 & (12, 4) & 65.47 $\pm$ 1.73 \\
        500 & (12, 4) & 67.58 $\pm$ 1.78 \\
        1000 & (16, 2) & \textbf{69.73} $\pm$ \textbf{1.57} \\
        \bottomrule
    \end{tabular}
    \caption{Test accuracy results for LC-SNNs on the EMNIST ``letters'' partition with varying numbers of $n_\textrm{filters}$ using the $2$-gram classification scheme. Each result is estimated from 10 independent train and test phases.}
    \label{tab:emnist_results}
\end{table}

For reference, grid search was performed over hyper-parameters of the baseline SNN, which obtained its best accuracy of 66.56\% $\pm$ 1.96\% with $n_\textrm{neurons}$ = 900. As with the LC-SNN, a single training epoch is used before the network is evaluated on the test partition. The best setting of LC-SNN hyper-parameters for $n_\textrm{filters} = 500$ has 648K trainable synapse weights, comparable to the best baseline network with 705.6K parameters, while achieving better performance. Again, as the number of neurons in the output layer of the baseline network is increased, the number of passes through the training data needed to converge in performance increases dramatically.
The average test confusion matrix for the best set of hyper-parameters for LC-SNNs with $n_\textrm{filters} = 1000$ is shown in Figure \ref{subfig:emnist_confusion}.

To our knowledge, this work is the first to report results on the training of spiking neural networks to classify the EMNIST ``letters'' partition. Although we focus on only the 26 capital letters of the English alphabet, it remains to be seen if this approach will scale to all 62 categories of the original dataset.

\subsection{Training convergence}
\label{ssec:convergence}

A desirable feature of the LC-SNN is the rapid convergence speed which the distributed sub-network learning enables. That is, since only neurons within the same sub-population of the output layer (as determined by unique receptive fields) compete to learn filters, as many neurons as there are receptive fields can fire and subsequently update their synapse parameters at once. This leads to many more parameter updates than was previously possible, and a faster convergence to the network's peak classification accuracy.

\begin{figure}[H]
    \centering
    \includegraphics[width=0.9\textwidth]{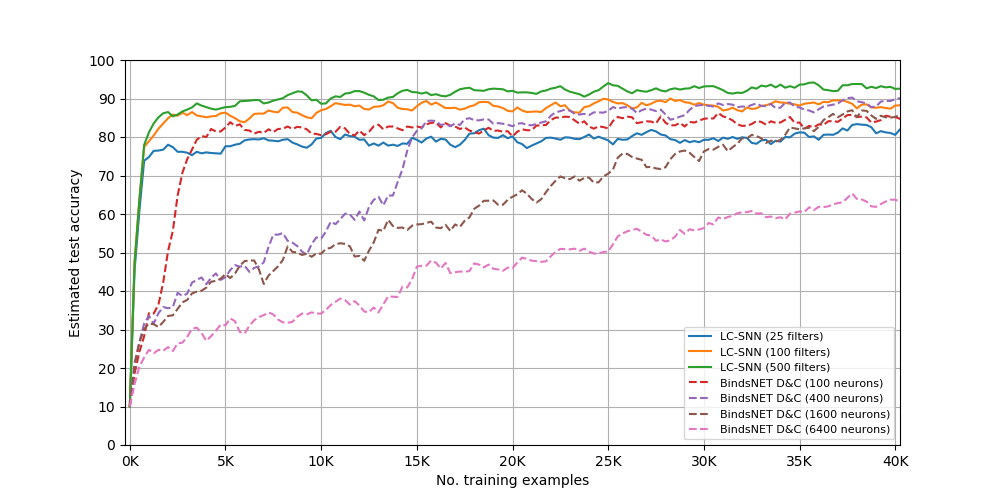}
    \caption{Training performance of our LC-SNN architecture and the fully-connected baseline SNN replicated in \texttt{BindsNET}  (denote D\&C in the legend). Shown are the estimated test accuracies on a single training phase (each using the same training data presentation order) evaluated at every 250 training examples.}
    \label{fig:convergence}
\end{figure}

Moreover, this convergence does not appear to depend on the size of the network; i.e., the number of locally-connected filters. On the other hand, the baseline SNN networks requires more training iterations as the size of the network increases in order to reach their best performance. This phenomenon is depicted in Figure \ref{fig:convergence}. Note that LC-SNNs often come with a few percentage points of their best accuracy within $\approx$ 5K training examples, whereas the baseline SNNs often require more than 60K examples to converge; e.g., for networks with 1,600 neurons, as demonstrated empirically in Section \ref{sssec:mnist}.

\section{Robustness tests}
\label{sec:robustness}

In the following, we perform the following robustness tests on a fixed, trained LC-SNN:

\begin{enumerate}
    \item Randomly deleting plastic synapses (from the input to the output layer) post-training.
    \item Randomly deleting neurons from the output layer (equivalent to deleting all incoming synapses to said neurons) post-training.
\end{enumerate}

We fix a single LC-SNN (with $k = 14, s = 2$, and $n_\textrm{filters} = 250$) and baseline SNN (with $n_\textrm{neurons} = 900$) which both achieved $\approx$ 93.5\% test accuracy on the MNIST dataset, which are used in all robustness tests. The filters of the LC-SNN are re-shaped to be two-dimensional and visualized in Figure \ref{subfig:lcsnn_filters}. Note that the baseline network required 3 passes through the training data in order to attain the same level of accuracy as the LC-SNN with only a single training epoch.

\subsection{Deleting synapses}
\label{ssec:synapses}

\begin{figure}[H]
    \centering
    \includegraphics[width=0.75\textwidth]{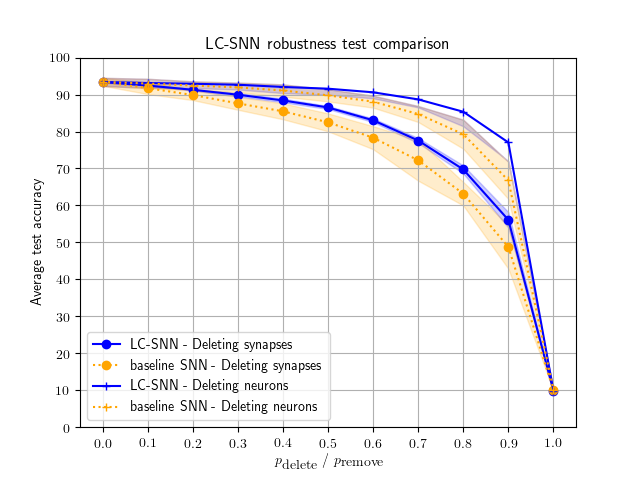}
    \caption{Robustness of LC-SNN vs. baseline SNN to random synapse and neuron deletion.}
    \label{fig:robustness}
\end{figure}

For each STDP-modifiable synapse in the aforementioned fixed LC-SNN, we delete it with probability $p_\textrm{delete}$, and run the test phase of the MNIST digits as normal. This gives us a sense of how performance degrades in the face of missing inputs; we run the test phase 5 times in order to get the average and standard deviation statistics of the performance result. The results are shown in Figure \ref{fig:robustness}.


As can be seen in the figure, the network without any synapses deleted maintains an accuracy of $\approx$ 93.5\%, while increasing $p_\textrm{delete}$ leads to a smooth degradation in performance. Interesting, with 50\% of synapses removed, nearly 90\% accuracy is maintained, and with 90\% of synapses removed, nearly 60\% accuracy is maintained. The baseline SNN drops off in performance at a faster rate than the locally connected network, possibly due to the distributed nature of the LC-SNN representation, as opposed to the fully-connected representation learned by the baseline.

\begin{figure}[H]
    \centering
    \begin{subfigure}[t]{0.3\textwidth}
        \includegraphics[width=\textwidth]{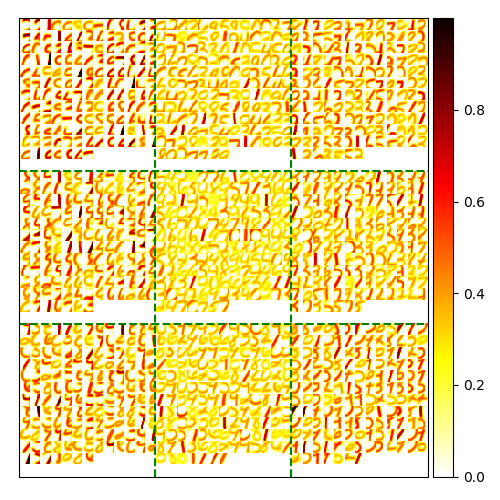}
        \caption{LC layer weights after 60K MNIST training iterations.}
        \label{subfig:lcsnn_filters}
    \end{subfigure}
    ~
    \begin{subfigure}[t]{0.3\textwidth}
        \includegraphics[width=\textwidth]{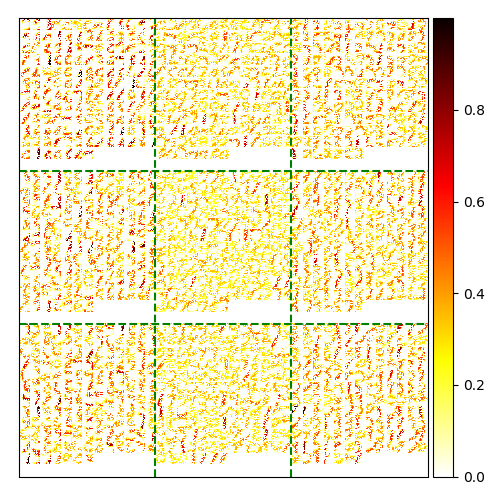}
        \caption{Synapses deleted with probability $p_\textrm{delete} = 0.5$.}
        \label{subfig:synapse_deletion}
    \end{subfigure}
    ~
    \begin{subfigure}[t]{0.3\textwidth}
        \includegraphics[width=\textwidth]{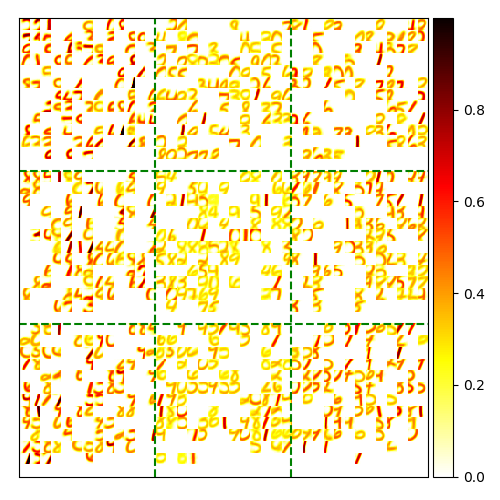}
        \caption{Neurons removed with probability $p_\textrm{remove} = 0.5$.}
        \label{subfig:neuron_removal}
    \end{subfigure}
    \caption{LC-SNN filters and random deletion of synapses / removal of neurons.}
    \label{fig:filter_robustness}
\end{figure}

An example deletion of synapses with $p_\textrm{delete} = 0.5$ is given in Figure \ref{subfig:synapse_deletion}.

\subsection{Deleting neurons}
\label{ssec:neurons}

Similar to the approach in Section \ref{ssec:synapses}, we consider neurons in the output layer of the LC-SNN, and remove each with probability $p_\textrm{remove}$. We can implement this simply by setting all incoming synapses to a ``removed'' neuron to 0, effectively cutting it off from any positive input activity. This experiment assesses the degree of redundancy encoded in individual neurons, as opposed to the previous experiment which assessed the importance of individual synapses. The MNIST test phase is run 5 times to obtain mean and standard deviation statistics of the test performance result, which are depicted Figure \ref{fig:robustness}.


From the figure, we conclude that the network filters encode a high degree of redundancy, as deleting large numbers of neurons results in relatively little reduction in accuracy. For example, deleting $90\%$ of neurons reduces accuracy down to just under $80\%$, reducing the number of filters down from 150 to approximately 15 per receptive field. The baseline SNN's accuracy drops at a faster rate, again possibly due to the difference in nature of the two architecture's learned representations. On the other hand, since the compared baseline SNN has fewer neurons, deleting a large percentage of them may have a more dramatic effect than for the LC-SNN.

An example removal of neurons with $p_\textrm{remove}$ is given in Figure \ref{subfig:neuron_removal}.


\section{Discussion}
\label{sec:discussion}

\subsection{Multi-thread decision algorithm}

Here we introduce an additional biologically-motivated method for deriving decisions from the LC-SNN spike trains. Previously results were obtained using the n-gram approach with very good accuracy. In the proposed multi-thread decision rule, we do not simply select the winning n-grams from all spike trains, but consider the location (patch) where the winning n-grams are coming from. In this way we contrast the temporal dimension of spike trains with their spatial distribution in multiple locations, representing multiple threads. Each of the patches form their own 2-gram dictionary and vote on the top N classes the ($1^{st}$ class gets the highest weight and the $N^{th}$ class gets the least weight). We also tested a multi-thread method of classifying based on the sum of spikes. Unlike the n-gram method, this method does not utilize the temporal aspect of the spikes. Based on the sum of spikes, patches associate each of their neuron with a different classes based on the training data of sum of spikes of the neuron. During classification, the patches vote on the top N classes. The overall decision is made by choosing the class that gets the most votes. This method of classification is much faster than the n-gram method as it does not require building n-gram and does not rely on the temporal aspect of the spikes.

The results with multi-thread algorithm applied to MNIST classification are summarized in Table \ref{tab:mnist_results}. Note that the multi-thread sum of spikes method where each patch vote for top 3 classes is very good and it gets close to the logistic regression accuracy of $94.39\%$.

\begin{figure}[H]
    \centering
    \includegraphics[width=0.55\textwidth]{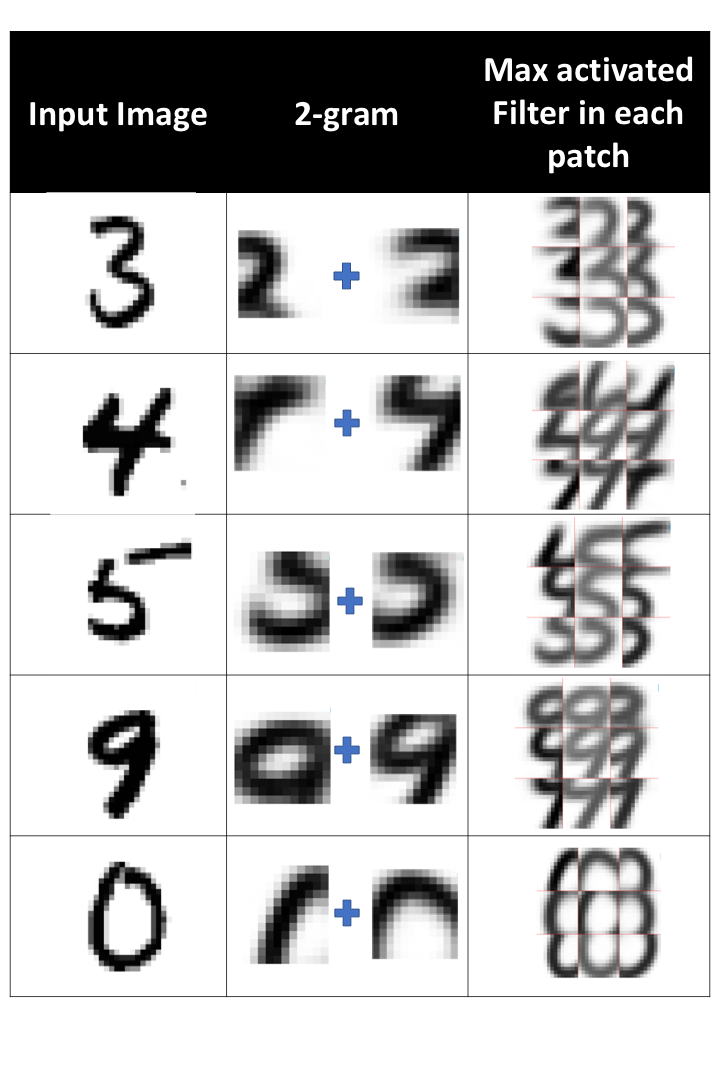}
    \caption{Illustration of the trained filters obtained with n-gram and multi-thread approaches for a few input representative digit classes.}
    \label{fig:multithread}
\end{figure}

It is important to point out that the multi-thread approach has strong biological motivation and it may provide insights on the operation of biological mechanisms of spike processing. This is illustrated by Fig. \ref{fig:multithread}, where the trained filters of the winning neurons are depicted for the single-thread 2-gram and multithread approaches. We can see that the multi-thread filters are in line with the expectation that a well-functioning processing system can efficiently extract information from various locations of the input data. This way the method has the promise of being scalable to increasing input data sizes.

\begin{table}[H]
    \centering
    \begin{tabular}{|c|c|}
        \toprule
        Method & Accuracy \\
        \midrule
       Single thread: 2-gram & $91.68$\\
        Multi-thread: 2-gram & $88.91$\\
        Multi-thread (Sum of spikes): 3 votes & $93.21$\\

        \bottomrule
    \end{tabular}
    \caption{Test accuracy results for LC-SNNs with different classification schemes. Each result is for $n_\textrm{filters} = 100$ and $(k,s) = (12, 4)$.}
    \label{tab:mnist_results}
\end{table}

\subsection{Concluding remarks}

We have demonstrated a new architectural building block, the locally-connected layer, for spiking neural networks and applied it to solve simple image recognition tasks. The layer allows two-layer networks to divide and conquer the input space by learning a distributed representation. Not only does this division of labor allow a reduction in the number of learnable parameters to achieve a certain quality of representation (as measured by classification accuracy), but it reduces the number of training examples required to converge to it. These LC-SNN properties have been empirically demonstrated on the MNIST and EMNIST datasets, where we have shown improved performance with shorter training times compared to a baseline, fully-connected SNN.

The locally-connected layer may be naturally included as a component in a multi-layer spiking neural network alongside with other feedforward, lateral, or recurrent connections. It encodes the inductive bias that features in the pre-synaptic space may be divided into possibly overlapping regions without loss of information, and for the added benefit of reduced parameter requirements and increased training convergence speed. Although the LC-SNN induces approximately 2-5 times more spikes during its operation than the considered baseline SNN, we argue that this is an acceptable tradeoff in light of the aforementioned convergence and accuracy properties. Indeed, we argue that the extreme sparsity of activity in the baseline networks let to a poor, centralized representation of the dataset akin to the notion of the gnostic (``grandmother'') cell in neuroscience \citep{doi:10.1177/107385802237175}. The LC-SNN layer may be implicated in neuromorphic chips and used to operate on event-based data, a domain in which spiking neural networks are known to shine \citep{10.3389/fnins.2018.00774}.

It remains to be seen whether methods of unsupervised learning with SNNs will be useful in learning representations of more complex datasets that, e.g., separate the categories of data for classification purposes. Whereas the optical character datasets considered in this work are relatively simple, to our knowledge, it has not yet been shown that SNNs trained with unsupervised, local learning rules can scale to the complexity of real-world image data. However, we have added a component to the SNN builder's toolbox that may invigorate further lines of work on adapting SNNs to more complex machine learning problems. Using it, we have shown that local learning rules and adaptive neurons combined with competitive inhibitory interactions may be applied to different SNN architectural primitives with similar benefits for feature learning.

\section*{Acknowledgements}

This work has been supported in part by Defense Advanced Research Project Agency Grant, DARPA/MTO HR0011-16-l-0006 and by National Science Foundation Grant NSF-CRCNS-DMS-13-11165.

\section{References}

\bibliographystyle{IEEEtran}
\bibliography{bibtex.bib}

\end{document}